\documentclass[review]{elsarticle}

\usepackage{graphicx}
\usepackage{multirow}
\usepackage{enumitem}
\usepackage{algcompatible}
\usepackage{float}
\usepackage[titlenumbered,linesnumbered,figure]{algorithm2e}

\pdfoutput=1
 
\journal{Medical Engineering \& Physics}

\begin{document}

\begin{frontmatter}

\title{Review of Fall Detection Techniques: A Data Availability Perspective}

\author[rvt]{Shehroz S. Khan\corref{cor1}}
\ead{s255khan@uwaterloo.ca}

\author[rvt]{Jesse Hoey}
\ead{jhoey@uwaterloo.ca}

\address[rvt]{David R. Cheriton School of Computer Science, University of Waterloo,\\200 University Ave W, Waterloo, ON N2L 3G1, Canada}
\cortext[cor1]{Corresponding author}

\begin{abstract}
A fall is an abnormal activity that occurs rarely; however, missing to identify falls can have serious health and safety implications on an individual. Due to the rarity of occurrence of falls, there may be insufficient or no training data available for them. Therefore, standard supervised machine learning methods may not be directly applied to handle this problem. In this paper, we present a taxonomy for the study of fall detection from the perspective of availability of fall data. The proposed taxonomy is independent of the type of sensors used and specific feature extraction/selection methods. The taxonomy identifies different categories of classification methods for the study of fall detection based on the availability of their data during training the classifiers. Then, we present a comprehensive literature review within those categories and identify the approach of treating a fall as an abnormal activity to be a plausible research direction. We conclude our paper by discussing several open research problems in the field and pointers for future research.
\end{abstract}


\end{frontmatter}


\section{Introduction}

Research in activity recognition has led to the successful realization of intelligent pervasive environments that can provide context, assistance, monitoring and analysis of a subject's activities that are usually backed up by advanced machine learning and vision algorithms \cite{Lara2013Survey,ke2013review,Incel2013Review}. However, a lot of this research is centred around developing techniques to identify normal Activities of Daily Living (ADL) either at an atomic level (e.g. walking, running, cycling) or at a higher level (e.g. preparing breakfast, washing hands). These techniques are generally applied to monitor a subject's movements, assess physical fitness and provide feedback. Though this research is useful, there can be scenarios where detection of abnormal activities become important, challenging and relevant. Missing out such abnormal activities can impose health and safety risks on an individual. Falling is one of the most common type of abnormal activity and the most studied \cite{Mubashir:2013:SFD,igual2013challenges}. In real life, most falls are caused by a sudden loss of balance due to an unexpected slip or trip, or loss of stability during movements such as turning, bending, or rising \cite{Robinovitch20041329}. 

Falls are the major cause of both fatal and non-fatal injury among people and create a hindrance in living independently. According to the report by {SMARTRISK} \cite{website:execsum2009}, in Canada in $2004$, falls constituted $25\%$ of all the unintentional injuries besides transport injuries or suicides, resulting in $2225$ deaths, $105,565$ hospitalizations and $883,676$ non-hospitalizations. The report also suggests that falls accounted for $50\%$ of all injuries that resulted in hospitalization, and was the leading cause of permanent partial disability ($47\%$) and total permanent disability ($50\%$). Falls were the leading cause of overall injury costs in Canada in $2004$, accounting for $\$6.2$ billion or $31\%$ of total costs besides other unintentional injuries. According to the WHO report \cite{website:whofall}, the frequency of falls increase with an increase in age and frailty. Older people living in nursing homes fall more often than those living in the community (around $30-50\%$) and $40\%$ of them experience recurrent falls \cite{website:whofall}. The reason is that most of the older adults living in the nursing homes are more frail and these facilities report fall incidences more accurately \cite{rubenstein1994falls}. According to the Public Health Agency of Canada \cite{website:phacfall}, older adults in Canada who were hospitalized due to a fall spent up to three weeks in the hospital, which is three times more than the average hospital stay among other age groups. 
Falls can impact a person both economically and psychologically. Experiencing a fall may lead to a fear of falling \cite{igual2013challenges}, which in turn can result in lack of mobility, social isolation, less productivity and can increase the risk of a fall. 

Falls occur infrequently and diversely. The rarity of occurrence of falls lead to a lack of sufficient data for them for training the classifiers \cite{khan2016classification}. More than one type of fall may also occur and their unexpectedness make it difficult to model them in advance. Collecting fall data can be cumbersome because it may require the person to actually undergo a real fall which may be harmful and unsafe. Alternatively, artificial fall data can be collected in controlled laboratory settings; however, that may not be the true representative of actual falls \cite{kangas2012comparison}. Analyzing artificially induced fall data can be good from the perspective of understanding and developing insights into falls as an activity but it does not simplify the difficult problem of detecting falls. Moreover, the classification models built with artificial falls are more likely to suffer from over-fitting on them and may poorly generalize on actual falls. The approaches that exclusively collect fall data still suffer from their limited quantity and ethics clearances. The Centers for Disease Control and Prevention, USA \cite{website:cdc} suggests that on an average, nursing home residents incur $2.6$ falls per person per year. If an experiment is to be set up to collect real falls and assuming an activity is monitored every second by a sensor, then we get around $31.55$ million normal activities per year in comparison to only $2.6$ falls. The data for real falls may be collected by running long-term experiments in nursing homes or private dwelling using wearable sensors and/or video camera. However, the fall data generated from such experiments will still be skewed towards normal activities \cite{Stone2015Fall} and it is difficult to develop generalizable classifiers to identify falls efficiently. In addition to very few or no labelled data, the diversity and types of falls further make it difficult to model them efficiently.  %

Most of the previous review papers on fall detection assume sufficient data for falls, and survey methods and techniques based on different types of sensors and specific feature extraction/selection methods. We argue that since falls are rare events, standard supervised machine learning methods may not be well-posed to identify them efficiently. Keeping this view in mind, we present a taxonomy for the study of fall detection methods that depends on the availability of fall data present during training the classifiers. This taxonomy is independent of the type of sensors used to capture human activities and specific type of feature extraction/selection methods. The taxonomy envisages the problem of fall detection from a real-world perspective where falls are not abundant but the normal ADL can be easily gleaned.

The rest of the paper is outlined as follows. In Section \ref{sec:survey}, we survey the existing review papers on fall detection, present their contributions, highlight their limitations and analyze their cumulative outcomes. In Section \ref{sec:taxonomy}, we present the proposed taxonomy for the study of fall detection methods based on the availability of fall data. Section \ref{sec:litrev} presents a comprehensive review of the current and significant research in the field of fall detection using the proposed taxonomy. In Section \ref{sec:abnormal}, we exclusively review research on methods that treat falls as abnormal activity or can be adapted for this task. We conclude the paper with open research questions and future direction in this field in Section \ref{sec:conc}.

\section{Survey of Existing Literature Review on Fall Detection}
\label{sec:survey}
In the last decade, several review papers on fall detection are published that discuss different aspects of the fall detection problem involving various classification techniques, types of sensors and specific feature engineering methods. In this section, we survey major review papers on fall detection and highlight their focus of research, contributions and limitations.

 Noury et al. \cite{noury2007fall} report a short review on fall detection with an emphasis on the physics behind a fall, methods used to detect a fall and evaluation criteria based on statistical analysis. They discuss several analytical methods to detect falls by incorporating thresholds on the velocity of sensor readings, detecting no-movements, intense inversion of the polarity of the acceleration vector resulting from impact shock and suggest that such methods will result in high false positive rates. They mention, since falls are rare, unsupervised machine learning techniques are likely to fail to identify the first fall event because it was not observed earlier. Supervised algorithms can only classify `known classes' on which they are trained and such techniques may label a rare activity, like a fall, as `Others' along with other activities e.g. to stumble, to slip etc. Yu \cite{Yu_2008} presents a survey on approaches and principles of fall detection for elderly patients. Yu first identifies the characteristics of falls from sleeping, sitting and standing and categorize fall detection methods based on wearable, computer vision and ambient devices. These approaches were further broken down into specific techniques such as based on motion analysis, posture analysis, proximity analysis, inactivity detection, body shape and $3$D head motion analysis. Yu mentions that a fall is a rare event and it is important to develop techniques to deal with such scenarios. Yu further addresses the need for generic fall detection algorithms and fusion of different sensors such as wearable and vision sensors for providing better fall detection solutions. Perry et al. \cite{Perry2009Survey} present a survey on real-time fall detection methods based on techniques that  measure  only the acceleration,  techniques that combine acceleration with  other  methods, and  techniques that  do not measure acceleration. They conclude that the  methods measuring acceleration are good at detecting falls. They also comment that placement of a sensor at the right position on the body can impact the accuracy of fall detection techniques.  
 
Hijaz et al. \cite{Hijaz2010Survey} present a survey on fall detection and monitoring ADL and categorize them into vision based, ambient-sensor based and kinematic-sensor based approaches. They identify kinematic-sensor based approaches that use accelerometer and/or gyroscopes as the best among them because of its cost effectiveness, portability, robustness and reliability. Mubashir et al. \cite{Mubashir:2013:SFD} present another survey on fall detection methods with an emphasis on different systems for fall detection and their underlying algorithms. They categorize fall detection approaches into three main categories: wearable device based, ambience device based and vision based. Within each category they review literature on approaches using accelerometer data, posture analysis, audio and video analysis, vibrational data, spatio-temporal analysis, change of shape or posture. They conclude that wearable and ambient devices are cheap and easy to install; however, vision based devices are more robust for detecting falls. Delahoz and Labrador \cite{Delahoz2014Survey} present a review of the state-of-the-art in fall detection and fall prevention systems along with qualitative comparisons among various studies. They categorize fall detection systems based on wearable devices and external sensors that includes vision based and ambient sensors. They also discuss general aspects of machine learning based fall detection systems such as feature extraction, feature construction and feature selection. They also summarize various classification algorithms such as Decision Trees, Naive Bayes, K-Nearest Neighbour and SVM; compare their time complexities and discuss strategies for model evaluation. They further discuss several design issues for fall detection and prevention systems including obtrusiveness, occlusion, multiple people in a scene, aging, privacy, computational costs, energy consumption, presence of noise, and defining appropriate thresholds. They also present a three-level taxonomy to describe the falling risks factors associated with a fall that includes physical, psychological and environmental factors and review several fall detection method in terms of design issues and other parameters. Schwickert et al. \cite{Schwickert2013} present a systematic review of fall detection techniques using wearable sensors. One of the major focus of their survey is to determine if the prior studies on fall detection use artificially recorded falls in a laboratory environment or natural falls in real-world circumstances. They observe that around $94\%$ studies use simulated falls. This is an important finding because it highlights the difficulty in obtaining real fall data due to their rarity. They also discuss that accelerometers along with other sensors such as gyroscopes, photo-diodes or barometric pressure sensors can help obtain better accuracy, and the placement of sensors on the body can be of importance in detecting falls. 

Zhang et al. \cite{zhang2015_petra_fall} present a survey of research papers that exclusively use vision sensors, where they introduce several public datasets on fall detection and categorize vision based techniques that uses single or multiple RGB cameras and 3D depth cameras. Pannurat et al. \cite{pannurat2014automatic} present another review for automatic fall detection by categorizing the existing platforms based on either wearable and ambient devices, and the classification methods are divided into rule-based and machine learning techniques. They present a detailed overview of different aspects of fall detection, including sensor types and placement, subject details, ADLs and fall protocols, extracting features, classification methods, and performance evaluation. They also compare several fall detection products based on size, weight, sensor type, battery, transmission range, features and comment on future trends in the area of fall detection. Igual et al. \cite{igual2013challenges} review $327$ research papers on fall detection and categorize them as either context-aware systems or wearable devices (including smartphones). The context-aware systems are further categorized as based on cameras, floor sensors, infrared sensors, microphones and pressure sensors. They point out that despite the use of many feature extraction and machine learning techniques adopted by researchers, there is no standardized context-aware technique widely accepted by the research community in this field. The major contributions of their survey are the identification of emerging trends, ensuing challenges and outstanding issues in the field of fall detection. They point out that the limited availability of real life fall data is one of the significant issue which could hinder the system performance. Ward et al. \cite{ward2012fall} present a review of fall detection methods from the perspective of use and application of technology designed to detect falls and alert for help from end-user and health and social care staff. They categorize the technologies for fall detection based on manually operated devices, body worn automatic alarm systems and devices that detect changes which may increase the risk of falling. They also comment that the users of fall detection technologies are concerned with privacy, lack of human contact, user friendliness and appropriate training, but they identify the importance and benefits of such systems within the community. In their study, health and social care staff appear less informed and less convinced on the benefits of fall detection technologies. There are several other survey papers on fall detection \cite{el2013fall, hegde2013technical, Patsadu2012Survey, spasova2014survey, chaudhuri2014fall, kulkarni2013review} that address similar ideas and issues already covered in this section.
 
In the past few years, smart phones have becomes very popular as they are non-invasive, easy to carry, work both indoors and outdoors and are equipped with sensors that are useful for activity recognition. There has been a considerable amount of research work done for general activity recognition and fall detection using smartphones. Luque et al. \cite{luque2014comparison} present a review of comparison and characterization of fall detection systems based on android smart phones. They mention that most of the techniques for fall detection based on smart phones either use machine learning (pattern matching) techniques or fixed threshold(s). They conduct experiments with simulated falls, compare them with several algorithms and observe that the accuracy of the accelerometer based techniques to identify falls depend strongly on the fall pattern. They also find difficulty in setting acceleration thresholds that allow achieving a good trade-off between false alarms and missed alarms. They further mention the hardware limitations of the memory and real-time processing capabilities of the smart phones that may not support complex fall detection algorithms. Another major problem raised is the rate of battery consumption when mobile application for continuous monitoring are used. Casilari et al. \cite{Casilari_2015} present another survey on the analysis of android based smart phones solutions for fall detection. They systematically classify and compare many algorithms from the literature taking into account different criteria such as the system architecture, sensors used, detection algorithms and the response in case of a false alarms. Their study emphasizes the analysis of the evaluation methods that are employed to assess the effectiveness of the detection process. 

\subsection{Analysis}
We observe several recurring themes that consistently appear among all the review papers we discussed previously:
\begin{itemize}
  \item There exists no standard methodology for fall detection in terms of type of sensors, feature engineering or machine learning techniques that supersedes other methods or perform consistently better than others. 
  \item It is noted in many of the above survey papers that techniques based on fixed thresholds on sensor readings, though simple to implement and computationally inexpensive, are very hard to generalize across different persons and does not provide a good trade-off between false positives and false negatives \cite{igual2013challenges}.
  \item Many of these survey papers reveal the complete lack of a reference framework, publicly available datasets and almost no access to real fall data to validate and compare to other methods.
  \item Most of the above discussed survey papers review research on fall detection that assume sufficient data for falls and or adequate prior knowledge and understanding of falls. A fall is a rare event that can occur in diverse ways \cite{Debard2012Camera}; therefore, collecting sufficient fall data is very difficult. A long term experiment is required to glean real falls; however, such an experiment may only result in very few samples for real falls \cite{Stone2015Fall,Debard2012Camera}.
  \item Some of the review papers we discussed above acknowledge the rarity of real falls and the difficulty in generalizing results obtained from artificial or simulated falls \cite{igual2013challenges,noury2007fall,Yu_2008,Schwickert2013}; however, they did not review techniques that may be capable of identifying falls when their training data is very limited or not present. 
  \item Since most of the above discussed research papers assume sufficient falls collected from laboratories, we could not get useful insights about setting up long term experiments for collecting real fall data. 
  \end{itemize} 
    
\section{Taxonomy for the Study of Fall Detection}
\label{sec:taxonomy}

Based on the inferences drawn from the recent review work on fall detection, we present a taxonomy for the study of fall detection methods that depends on the availability of training data for falls (see Figure \ref{fig:taxonomy}). This taxonomy is independent of the type of sensors to capture human motions and specific feature engineering methods employed to tackle this problem. The proposed taxonomy focuses on investigating classification methods based on availability of data for falls. Specific sensors (e.g. wearable, vision, ambient etc) can be used for the task for fall detection. Similarly, specific feature extraction and selection methods can be used based on the type of data captured with the sensors.\footnote{Interested readers can consult the review papers discussed in Section \ref{sec:survey} for sensor specific fall detection techniques along with dedicated feature extraction methods.}. 
The taxonomy has two high level categories: 
\begin{enumerate}[label=(\Roman*)]
  \item Sufficient training data for falls
  \item Insufficient or no training data for falls
\end{enumerate}

The category \ref{cat1} of the taxonomy shows the case when sufficient data for falls is available for training the classifiers. Even though real falls occur rarely, sufficient amount of fall data can be collected using simulated falls. In this category, due to the presence of sufficient falls and normal ADL data, different algorithms based on supervised machine learning, threshold(s) and one-class classifiers (trained only on sufficient fall data) can be employed. This category represents an optimistic view of the difficult problem of fall detection. 
However, even for these methods, the generalization of fall detection results across different persons is challenging \cite{medrano2016effect}, mostly because of significant variations in the properties between real and simulated falls \cite{klenk2011comparison}. 
In a real world scenario, we may expect either too few falls or none to begin with due to their rarity and difficulties in the data collection process. In these highly skewed data scenarios, heuristic and traditional supervised classification algorithms may not work as desired and other classification frameworks based on over/under-sampling, semi-supervised learning, cost-sensitive learning, outlier/anomaly detection and One-Class Classification (OCC) are needed; these techniques are mentioned in category \ref{cat2} of the taxonomy. 
Both the categories \ref{cat1} and \ref{cat2} assume sufficient training data for normal ADL and only differ in the amount of fall data available during training; however, both of these categories give rise to different types of approaches for fall detection. The approaches in category \ref{cat1} attempt to detect a fall directly given their training data, whereas the approaches in category \ref{cat2} either manipulate little available fall data or try to indirectly detect a fall as an abnormal activity given (almost) no training data for them. 

\begin{figure}[H]  
\centering
  \begin{algorithm}[H]
    \begin{enumerate}[label=(\Roman*)]

      \item \textbf{Sufficient data for falls} \label{cat1}

      \begin{enumerate}
        \item Apply supervised machine learning techniques.\label{1a}
        \item Apply threshold based techniques to detect falls from normal activities.\label{1b}
        \item Apply OCC techniques to filter out normal activities as outliers.\label{1c}
      \end{enumerate}

      \item \textbf{Insufficient data for falls} \label{cat2}

      \begin{enumerate}
        \item If some fall data is available, apply over/under-sampling techniques.\label{2a}
        \item If some fall data is available along with a lot of unlabelled data, apply semi-supervised techniques.\label{2b}
        \item If some fall data is available, apply cost sensitive classification algorithms \label{2c}
        \item If no fall data is available, apply outlier / anomaly / density based detection techniques.\label{2d}
        \item If no fall data is available, apply OCC techniques.\label{2e}
      \end{enumerate}%
    \end{enumerate}%
  \end{algorithm}%
\caption{Taxonomy for the study of fall detection methods.}
\label{fig:taxonomy}
\end{figure}

The OCC approaches for fall detection appear in both category (I) and (II); however their underlying principle to detect falls is different. The category (I) assumes sufficient amount of fall data; therefore, an OCC can be trained on only fall data and can be used to identify a test sample as a fall or not-fall (or normal activity in our case, see Taxonomy \ref{1c}). Similarly, category (II) assumes insufficient or no training data for falls; therefore an OCC can be trained on the sufficiently available normal data. This classifier can be used to identify a test sample as a normal or not-normal activity (or fall in our case, see Taxonomy \ref{2e}). 

The techniques mentioned in category \ref{cat1} that directly attempt to detect falls are mostly discriminative and may use domain knowledge about the falling event (e.g. sudden change in acceleration or its short duration) \cite{Debard2012Camera} . 
There are three major drawback of these approaches (i) they cannot handle a realistic scenario when there is no training data for falls and the classifier will always make mistake in prediction because falls class was never observed, (ii) at the best, they can classify falls as members of some `Others' category; however, there should be some samples available for that category as well, and (iii) a lot of time and effort may be spent on labelling the data.

The techniques mentioned in category \ref{cat2} can be discriminative if they use some falls or generative if they detect falls as abnormal activity as an indirect evidence on the occurrence of falls \cite{Debard2012Camera}. This evidence may include prolonged inactivity, unusual locations, sudden change from normal behaviour and unknown or unseen behaviours. However, a major challenge is such techniques is to identify the first occurrence of a fall. This primarily depends on the definition of ``what is a normal behaviour?'', which needs to be defined carefully as it can vary across different persons, specially for different age groups. These techniques only need to learn the normal behaviour; therefore, the inherent data imbalance between normal activities and falls is not an issue because they do not need samples for falls (or their different types) during training of the classifier. However, if the normal behaviour is not properly learned, these systems can result in large amount of false alarms because any slight variation from the normal behaviour would be classified as a fall. Finding an optimal threshold that can minimize both false alarms and missed alarms in these techniques is very challenging \cite{DBLP:journals/tkde/YinYP08}. It is important to note that every abnormal behaviour or deviation from the normal behaviour does not imply the occurrence of a fall incident. As a result of these problems, such techniques require a lot of training data to effectively capture the normal behaviour over a long duration. The techniques that treat a fall as an abnormal activity does not require labelled data (models are only trained for normal activities) and a lot of time and effort can be saved.

In the next sections, we review the literature based on the taxonomy for fall detection described above. We will not further discuss many supervised methods for fall detection and interested readers may find those references in the survey papers on fall detection discussed in Section \ref{sec:survey}. In the literature review, we present algorithms that 
\begin{itemize}
\item Work with different types of sensors.
\item Use a variety of machine learning algorithms, especially those that are known to model temporal and sequential data (both for falls and ADL).
\item Can train models using only fall data, and
\item May work with a small amount of (or none) training data for falls using sampling techniques, cost-sensitive learning and semi-supervised learning approaches.
\end{itemize}

\section{Literature Review}
\label{sec:litrev}

\subsection{Sufficient data for falls}
Most of the research work in fall detection deals with applying supervised classification methods. In this paper, we are not reviewing such techniques and interested readers may refer to the papers cited in Section \ref{sec:survey}. Several research works in fall detection are based on thresholding techniques \cite{Bourke200884,dai2010perfalld,Jung2015Inertial,Chen1617246,Abba1101:Recognition,Lan:2009:SAF,DBLP:conf/bsn/LiSHBLZ09,Jung2015Inertial}, wherein raw or processed sensor data is compared against a single or multiple pre-defined thresholds to detect a fall (see Taxonomy \ref{1b} in Figure \ref{fig:taxonomy}). Most of these techniques need training data for falls and employ either domain knowledge or data analysis techniques to compute threshold(s) for their identification. The problem with thresholding techniques for fall detection is that it is very difficult to adapt thresholds to new types of falls and generalize them across different people \cite{bagala2012evaluation,igual2013challenges}. More information on these methods can be found in the survey papers discussed in Section \ref{sec:survey}.

\subsubsection{One-Class Classification}
If sufficient fall data are available, one-class classifiers can be trained on falls to reject normal activities (see Taxonomy \ref{1c} in Figure \ref{fig:taxonomy}). However, in realistic settings such a  strategy is highly unlikely because the availability of sufficient real fall data is difficult . Zhang et al. \cite{hunag:li:irwin:2006} train  OSVM from falls and outliers from non-fall ADL and show that falls can be detected effectively. Yu et al. \cite{Miao:Naqvi:2011} propose to train Fuzzy OSVM on fall activity captured using video cameras and to tune parameters using fall and some non-fall activities. Their method assigns fuzzy membership to different training samples to reflect their importance during classification and is shown to perform better than OSVM. Yu et al. \cite{Yu2012One} introduce a video-based fall detection system for elderly people. They extract several video features and apply OCC techniques to determine whether the new instances lie in the `fall region' or outside it to distinguish a fall from other activities such as walking, sitting, standing, crouching or lying. They test four OCC methods; $K$-center, $K$ nearest neighbour, OSVM and single class minimax probability machine (SCMPM) and find that SCMPM achieves the overall best performance among them. Han et al. \cite{han2014wifall} propose to use wireless signal propagation by employing the time variability and special diversity of Channel State Information as the indicator of human activities. Firstly a local outlier factor algorithm \cite{breunig2000lof} is used to filter out dynamic activities such as walking, sitting, standing up and falling and then an OSVM is trained on fall activities to distinguish it from other normal activities. 

\subsection{Insufficient data for falls}
\label{sec:insufficient}
\subsubsection{Sampling Techniques}
In some situations, there may be few real falls available during training phase (see Taxonomy \ref{2a} in Figure \ref{fig:taxonomy}). In these cases, either the minority `fall' class be over-sampled or the majority normal activity class be under-samples to train supervised classifiers. Stone and Skubic \cite{Stone2015Fall} present a two-stage fall detection system. In the first stage, a person's vertical state is characterized in individual depth image frames followed by an ensemble of decision trees to compute a confidence on the occurrence of a fall. The data they collected has a fall ratio w.r.t. normal activities of $1:400$. They under-sample the normal activities s.t. the ratio is reduced to $1:40$ and create a decision tree ensemble. The activities studied in their analysis are standing, sitting, and lying down positions, near (within $4$ m) versus far fall locations, and occluded versus not occluded fallers and report better results in comparison to the state-of-the-art methods. Debard et al. \cite{Debard2012Camera} use a weighted SVM to handle the imbalance in the dataset obtained for real world falls and normal activities from camera. The weights were computed using cross-validation and a grid search maximizing the area under the curve of a Receiver Operating Characteristic (ROC) curve. However, oversampling techniques may suffer from over-fitting if too many artificial data points are generated that do not represent actual falls, or under-fitting if the normal class is too much under-sampled to match the size of falls class.

\subsubsection{Semi-Supervised Techniques}
Labelling of activities manually can be very cumbersome, exhausting and time-consuming. In most of the cases, there is plenty of unlabelled data along with some labelled data; such a case calls for the implementation of semi-supervised methods for fall detection (see Taxonomy \ref{2b} in Figure \ref{fig:taxonomy}). Liu et al. \cite{Liu2008Vision} present a vision based semi-supervised learning mechanism for detecting falls and other ADL to overcome the exhaustive labelling of human activities. They use spatial field constraint energy to assist SVM-based activity decision with a Bayesian inference model, followed by semi-supervised step to retrain the classifier by automatic annotating the activities with the highest confidence. However, their method is sensitive to changes in environment, and needs to be retrained in new situations. Fahmi et al. \cite{Fahmi2012Semi} present a semi-supervised fall detection method using smartphones by first training a supervised algorithm using decision trees, then using fall profiles to develop a semi-supervised algorithm based on multiple thresholds. Medrano et al. \cite{Medrano2014Personalizable} present a nearest neighbour based semi-supervised fall detection method for smartphones that can be personalized and updated easily as a new user records new ADL and the system is retrained on the fly. Makantasis et  al. \cite{Makantasis20153D} present a $3$D semi-supervised fall detection system that uses a monocular camera and uses an expert to refine an initially created small subset of labelled activity samples. Yang et al. \cite{Yang2016} present a semi-supervised approach to detect near-miss falls for ironworkers. They emphasize that collecting and labelling data for near-miss falls is very difficult and use abnormal signals as stand-ins for near misses. They collect data through IMU device and train OSVM on normal walking patterns; any observation that lies beyond the classification boundary is considered as a near-fall.

\subsubsection{Cost Sensitive Classification and Decision Theory}
\label{sec:costsensitive}

Learning from imbalanced data pertains to situations where data from one class is available in abundance, but the data from the other class is rare, difficult to collect or not readily available. The problem of fall detection is a good example for imbalanced learning because the data for normal activities are easy to collect and sufficiently available in comparison to falls. Traditional supervised algorithms expect balanced datasets with equal misclassification costs for different classes. However, these algorithms fail to represent characteristics of the data and provide unfavourable accuracies when presented with imbalanced dataset \cite{he2009Learning}, and their predictions may be dominated by the majority class \cite{klement2011classifying}.  To handle imbalanced classes, cost-sensitive classification is performed \cite{he2009Learning}, where the main idea is to treat different costs in a classification problem differently. This can be done by either presenting different cost matrix to a cost-insensitive classifier or by changing the inner workings of a classification algorithm such that it uses a cost function to build a cost-sensitive classifier. Let us take the case of fall detection, where the dataset is mostly imbalanced in favour of normal activities. Identifying a rare activity (such as a fall) is important from the health and safety perspective, but the cost of a false alarm and missing to report a fall should be very different and must not be treated equally. Therefore, when some data for falls is present for analysis along with abundant data for normal activities, techniques based on cost-sensitive classification can be used for fall detection (see Taxonomy \ref{2c} in Figure \ref{fig:taxonomy}). However, the incorporation of cost of classification while reporting or not-reporting a fall is absent in most of the studies on fall detection. It is important to note that the cost of errors in this problem should be domain-specific and not vary across different datasets. However, such costs are mostly unknown and hard to compute \cite{khan2016classification}. 
Huang et al. \cite{huang2011enhanced} perform cost-sensitive analysis for fall detection using Bayesian minimum risk and the Neyman-Pearson method. They vary the ratio of the cost of a missed alarm to a false alarm to find an optimal region of operation using the ROC curve. On the contrary, this ratio is generally fixed and must not depend on the dataset. The technique presented in the paper to estimate cost ratio can overfit the dataset without providing any intuitive interpretation about it. 
However, traditional approaches for cost-sensitive learning for imbalanced datasets may not be directly applicable in the OCC problems. The reason is that in OCC case, the data for one of the class is absent during training; therefore, the probability estimates for the unseen or abnormal class is hard to compute and estimating the costs of errors in such cases are even harder. 

This problem can also be stated from a decision-theoretic perspective because costs and utilities are related concepts. Decision theory pertains to rational decision making by agents and can be employed to compute the expected cost of (not)reporting a fall. 
There is very sparse literature on decision-theoretic methods for identification of outliers or rare events/activities. Decision-theoretic approaches have been applied in some tasks, including detecting anomalies in internet path \cite{fida2011internetpath}, intrusion detection \cite{maloof2006machine}, fraud detection \cite{torgo2011utility} and fault detection in wireless sensor networks \cite{nandi2014hypothesis}. For the fall detection application, Khan and Hoey \cite{khan2016dtfall} present a decision-theoretic framework to report unseen falls given an arbitrary amount (possibly zero) of training data for falls, and given little or no information about the costs associated with falls. They present a novel method to parameterize unseen falls to accommodate situations when no fall data is available during training. Their results show that the difference in the cost of between a reported and non-reported fall is not that useful. However, the central question to derive exactly those values of cost remain open.

Table \ref{tab:study1} shows the summary of methods discussed in Section \ref{sec:insufficient} to detect falls with insufficient data that includes sampling techniques, semi-supervised techniques and cost-sensitive methods. We observe that different machine learning methods are used to handle this problem for both wearable and vision sensors and many of the methods incorporate domain knowledge.

\begin{table}[!ht]
\centering
\begin{tabular}
{|l|l|l|}\hline
\textbf{Type of Sensor} & \textbf{Authors} & \textbf{Technique Used} \\ \hline
\multirow{4}{*}{Wearable Device}            & Fahmi et al. \cite{Fahmi2012Semi} & Decision Trees, thresholds \\ \cline{2-3} 
                                            & Medrano et al. \cite{Medrano2014Personalizable} & Semi-supervised classification \\ \cline{2-3}
                                            & Yang et al. \cite{Yang2016} & OSVM \\ \cline{2-3}
                                            & Huang et al. \cite{huang2011enhanced} & Sensitivity Optimization \\ \cline{2-3}
                                            & Khan and Hoey \cite{khan2016dtfall} & GMM, Decision Theory \\ \hline
                                            
\multirow{4}{*}{Video}                      & Stone and Skubic \cite{Stone2015Fall} & Random Forest, Under-sampling \\ \cline{2-3}
                                            & Debard et al. \cite{Debard2012Camera} & Weighted-SVM \\ \cline{2-3}
                                            & Liu et al. \cite{Liu2008Vision}  & SVM, Bayesian Inference \\ \cline{2-3}
                                            & Makantasis et  al. \cite{Makantasis20153D} & Visual cues \\ \hline
\end{tabular}
   \caption{Summary of Studies on Detecting Falls with Insufficient training data}
   \label{tab:study1}
  \end{table}

\section{Detecting Falls in the Absence of their Training Data}
\label{sec:abnormal}

In a realistic setting, due to the lack of availability of sufficient data for falls and the lack of knowledge and understanding of what those falls might be, approaches based on outlier/anomaly detection (see Taxonomy \ref{2d} in Figure \ref{fig:taxonomy}) and one-class classification (see Taxonomy \ref{2e} in Figure \ref{fig:taxonomy}) can be employed for detecting falls. These techniques may not identify falls directly because fall samples are not available to the classifiers during the training phase; however, they can identify falls indirectly as an abnormal (or not-normal) activity. In these approaches, deviations from normal behaviour are flagged as an abnormal activity. Falling is one of the most common abnormal activities; therefore, such techniques can be adapted for fall detection. However, the concept of normal activities must be clearly defined because every abnormal activity may not be a fall. If the normal activities data is not sufficient, then such techniques may produce excessive false alarms. Recent research works \cite{medrano2014detecting,micucci2015falls,khan2014iwaal,Khan:2012:TDU:2370216.2370444} show that falls can be identified without seeing them in the past or specific domain knowledge about them. Below, we review techniques that build classification models using only the normal activities data and treat fall as an anomaly.

\subsection{Studies on Detecting Falls as Anomalies}
Zhou et al.\cite{Zhou2012An} present a method to detect falls using transitions between the activities as a cue to model falls. They train supervised classification methods using normal activities collected from a mobile device, then extract transitions among these activities and use them to train an OSVM and show that it performs better than an OSVM trained with only normal activities. Yu et al. \cite{DBLP:journals/titb/YuYRNWC13} present an online OSVM learning algorithm to detect falls captured through a single video source. They extract three types of features: ellipse, shape-structure and position features to build the normal model by an online OSVM which can be updated to new emerging postures. Additional rules were added to the system to report fewer false alarms and to improve fall detection performance. Popescu \cite{Popescu2009} presents a fall detection technique that uses acoustic signals of normal activities for training and detecting fall sounds from it. They train OSVM, one-class nearest neighbour approach  (OCNN) and One-class Gaussian Mixture Model (GMM)  (that uses a threshold) to train models on normal acoustic signals and find that OSVM performs the best. However, it is outperformed by its supervised counterpart. Khan et al. \cite{SalmanKhan2015199} propose an unsupervised acoustic fall detection system with interference suppression that makes use of the features extracted from the normal sound samples, and constructs an OSVM model to distinguish falls from non-falls. They show that in comparison to Popescu \cite{Popescu2009}, their interference suppression technique makes the fall detection system less sensitive to interferences by using only two microphones. Medrano et al. \cite{medrano2014detecting} propose to identify falls using a smartphone as a novelty from the normal activities and find that OCNN performs better than OSVM but is outperformed by supervised SVM. Micucci et al. \cite{micucci2015falls} evaluate fall detection methods that do not require fall data during training on different datasets collected using smartphone accelerometers. Their results show that in most of the cases, one-class K-nearest neighbour approach performs better or equivalent to the supervised SVM and KNN classification approaches that require data for both the normal and fall activities. The main contribution of this study is the finding that to design an effective fall detection method, prior understanding of falls patterns is not necessary.
Khan et al \cite{khan2014iwaal} present two `X-Factor' HMM approaches that are like normal HMMs, but have inflated output covariances that can be used as alternative models to estimate the parameters of unseen falls. The inflated covariance parameter can be estimated using a cross-validation on the set of `outliers' in the normal data that serve as proxies for the (unseen) fall data that allows the XHMMs to be learned from only normal activity data. Their results show high detection rates for falls on two activity recognition datasets, albeit with an increase in the number of false alarms. In an extended work, Khan et al. \cite{khan2015detecting} show empirically that `X-Factor' approach with no fall data can perform better than supervised classification data with few falls. When the number of fall data increases, the supervised classifiers' performance improve, but collecting sufficient fall data can take lot of time (in years), effort, resources and extensive setup to conduct such experiment. 
Taghvei and Kosuge \cite{taghvaei2014hmm} propose a method for a real-time visual state classification of a user with a walking support system. They extract visual features using principal component analysis and use an HMM  for identifying real-time falls and state-recognition. In their experiment, an HMM  is trained on walking activity and a threshold is separately calculated using a safe distance from the minimum probability distribution value of normal walking, which is adjusted with a constant. If the log-likelihood of a new activity is more than the calculated threshold, then it is identified as either a fall, sit or stand. A multi-class step is performed on top of the one-class classifier to identify one of the states of the HMM  that corresponds to different activities including falls. A major drawback of this method is that the threshold is chosen only based on one activity, but sitting and standing are also normal activities. Taghvaei et al. \cite{Taghvaei2012Comparative} present another fall detection method from a walker that uses features obtained from a depth camera and uses one-class GMM to identify non-walking states and a continuous HMM  for identifying different types of falls. The threshold for one-class GMM  that represents normal activity is set experimentally. However, it is difficult to generalize this type of threshold across different people. Moreover, the threshold for a one-class GMM  is only obtained using walking activity and it classifies sitting and falls as outlier activities. Rougier et al. \cite{Rougier2011} present a fall detection method using video feeds by tracking the person's silhouette and performing shape analysis. They use GMM  to distinguish falls from the normal data by manually setting a threshold on the log-likelihood. The above survey is presented in tabular form in Table \ref{tab:study2}.

\begin{table}[!ht]
\centering
\begin{tabular}
{|l|l|l|}\hline
\textbf{Type of Sensor} & \textbf{Authors} & \textbf{Technique Used} \\ \hline
\multirow{4}{*}{Wearable Device}            & Zhou et al.\cite{Zhou2012An} & OSVM \\ \cline{2-3} 
                                            & Medrano et al. \cite{medrano2014detecting}  & OCNN, OSVM \\ \cline{2-3} 
                                            & Micucci et al. \cite{micucci2015falls} & OCNN \\ \cline{2-3}
                                            & Khan et al \cite{khan2014iwaal,khan2015detecting}  & HMM \\ \hline
                                         
\multirow{4}{*}{Video}                      & Yu et al. \cite{DBLP:journals/titb/YuYRNWC13} & OSVM \\ \cline{2-3}
                                            & Taghvei and Kosuge \cite{taghvaei2014hmm} & HMM \\ \cline{2-3}
                                            &  Taghvaei et al. \cite{Taghvaei2012Comparative} & HMM \\ \cline{2-3}
                                            & Rougier et al. \cite{Rougier2011} & GMM \\ \hline
                                         
\multirow{2}{*}{Acoustic}                   & Popescu \cite{Popescu2009} & OSVM, OCNN, GMM \\ \cline{2-3}
                                            & Khan et al. \cite{SalmanKhan2015199}  & OSVM \\ \hline
\end{tabular}
   \caption{Summary of Studies on Detecting Falls as Anomalies}
   \label{tab:study2}
  \end{table}

\subsection{Studies on Anomaly Detection not Specifically Focused on Fall Detection}
We discussed in Section \ref{sec:taxonomy} that fall detection methods in category \ref{cat2} (see Taxonomy \ref{2d} and \ref{2e} in Figure \ref{fig:taxonomy}) do not detect falls directly because either the training data for falls is insufficient or absent during training. In such scenario, a plausible approach is to identify a fall as an abnormal activity (or an anomaly), which we have discussed in the above section. Alternatively, several abnormal activity detection methods that are not specific for fall detection can also be adapted to detect a fall as one of the abnormal activity or an anomaly.
 Now, we review several abnormal activity recognition techniques that capture data from either vision-based or wearable sensors; only require training data from normal activities to learn the models and are able to identify abnormal activities. These methods can very well be adapted for the task of fall detection.

\subsubsection{Vision based}
\label{sec:vision_based}
Several approaches have been proposed for abnormal activity recognition using the computer vision sensors. Xiang and Gong \cite{DBLP:conf/iccv/XiangG05a} propose a Dynamic Bayesian Network approach to model each normal video pattern and use a threshold to detect abnormal activity. This approach is simple; however, choosing a threshold remains challenging. Duong et al. \cite{DBLP:journals/ai/DuongPBV09} model the duration of activities using a Coxian distribution \cite{DBLP:books/daglib/0018540} and consider a hierarchy of activities to propose a Switching Hidden Semi-Markov Model  and show its application to activity segmentation and abnormality detection in smart environments. Zhang et al. \cite{DBLP:conf/cvpr/ZhangGBM05} propose a semi-supervised adapted HMM  framework for audio-visual data streams which comprises of supervised learning of normal data and unsupervised learning of unusual events using Bayesian adaptation. Their method has an iterative structure, where each iteration corresponds to a new detected unusual event. However, it is not clear from their work how many iterations are needed to terminate the process of outlier detection. Their model assumes that the normal activities data contains unusual events and guarantees one outlier per iteration that could result in high false alarm rate. Jiang et al. \cite{ieee:Jiang:4379786} mention that the HMM  trained on small number of samples can overfit and propose a dynamic hierarchical clustering method based on a multi-sample based similarity measure. Their method starts with clustering the data in a few groups; the groups containing large number of samples are treated as normal patterns and HMMs are learned for each of them. This is followed by an iterative procedure of merging similar clusters (re-classifying) and re-training the remaining HMMs until no more merging occurs. An abnormal event is identified if its maximum log-likelihood from all normal events is below a threshold. They show their results on real surveillance video and point out that following the proposed method, the initial training and clustering errors due to overfitting will be sequentially corrected in later steps. Pruteanu-Malinici and Carin \cite{DBLP:journals/tip/Pruteanu-MaliniciC08} propose infinite HMM  modelling to train normal video sequences; unusual events are detected if a low likelihood is observed. 
Zhang et al. \cite{DBLP:conf/pakdd/ZhangLGH09} propose an abnormal event detection algorithm from video sequences using a three-phased approach. First, they build a set of weak classifiers using Hierarchical Dirichlet Process Hidden Markov Model (HDP-HMM) and then use ensemble learning to identify abnormal events. Finally, they extract abnormal events from the normal ones in an unsupervised manner to reduce the false positive rates. Hu et al. \cite{DBLP:conf/ijcai/HuZYZY09} propose a refinement of the HDP-HMM  method by incorporating Fisher Kernel into OSVM instead of ensemble learning and using sensor data instead of video data that can be discrete or continuous. The advantage of their method relies on using the HDP-HMM  models that can decide on the appropriate number of states of the underlying HMM  automatically. Antonakaki et al. \cite{DBLP:journals/sigpro/AntonakakiKP09} combine the use of HMM  and OSVM to detect abnormal human behaviour using multiple cameras. They treat short term behaviour classification and trajectory classification as separate classification problems by providing different set of features to both HMM  and OSVM. Two feature vectors are computed per instance to capture short term behaviour and trajectory information and fed to OSVM and HMM. Utilizing these two views of the data, they fuse the output of both the classifiers (logical OR) to identify abnormal activities. 
Matilainen et al. \cite{DBLP:conf/acivs/MatilainenBS09} present an unusual activity recognition method in noisy environments that uses a body part segmentation (BPS) algorithm \cite{DBLP:conf/icmcs/BarnardMH08}, which gives an estimation of similarity between the current pose to the poses in the training data. The normal activities they considered are are walking and sitting down and everything else is considered abnormal. The BPS algorithm uses an HMM  and a GMM  which are trained through synthetic data created by motion capture. They use three sequences containing walking and falling over as the training set to find a statistically optimal threshold for unusual poses. They also propose to use a majority voting over large number of consecutive decisions for the actions that spans over a period of time and to mitigate the effect of single frames with incorrect decisions that helps in reducing false positive rates. However, the approach is based on carefully chosen optimal thresholds and training on a minimalist set of synthetic normal activities, which render this approach not very useful\footnote{A detailed review on several other types of video-based abnormality detection methods is provided by Popoola and Wang \cite{6129539}}. 

Mahajan et al. \cite{mahajan04} propose an activity recognition framework based on multi-layered finite state machines (FSM) built on top of a low level image processing module for spatio-temporal detection and limited object identification. The FSM learns the model for normal activities over a period of time in an unsupervised manner and can identify deviant activities as abnormal. Parisi and Wermter \cite{Parisi2013Hierarchical} propose a hierarchical Self Organizing Maps (SOM) based architecture for the detection of novel human behaviour in indoor environments by learning normal activities in an unsupervised manner using SOM and report novel behaviour as abnormal. To handle tracking errors, a first SOM is used to remove outliers from the training motion vectors. The pre-processed motion vectors are encoded by three types of descriptors -- trajectories, body features and directions. A separate SOM is then trained for each type of descriptor. If a new observation deviates from the normal behaviour it is flagged as abnormal. Two thresholds are empirically defined to remove outliers from the first SOM and detect abnormal behaviour from the other three SOMs. 

\subsubsection{Wearable Sensor Based}
\label{sec:sensor_based}
Several recent works focus on using sensor networks to detect abnormal activities. Yin et al. \cite{DBLP:journals/tkde/YinYP08} propose a two-stage abnormal activity detection method in which an OSVM is first trained on normal activities and the  abnormal activities are filtered out and passed on to a kernel nonlinear regression routine to derive abnormal activity models from a general normal activity model in an unsupervised manner. The method iteratively detects different types of abnormal activities based on a threshold. They claim that this method provides a good trade-off between false alarms and abnormal activity detection without collecting and labelling the abnormal data. The data is collected by using wearable sensors attached to a user and abnormal instances were collected by simulating `falls' and `slipping' in different positions. 
Quinn et al. \cite{quinn} present a general framework of Switched Linear Dynamical Systems (SLDS) for condition monitoring of a premature baby receiving intensive care. They introduce the `X-factor' to deal with unmodelled variation from the normal events that may not have been seen previously. The general principle to identify an unusual event is to vary the covariance of the model of normal events to determine the interval with the highest likelihood where events can be classified as `not normal'. To model dynamic detection of abnormal events, they add a new factor to the existing SLDS model by inflating the system noise covariance of the normal dynamics. The sensor data is collected using various probes connected to each baby. The computation of the factor related to increasing the covariance remains challenging and is critical in this application. Cook et al. \cite{cook2012act} present a method for activity discovery (AD) for smart-homes to identify behavioural patterns that do no belong to the pre-defined classes. Their algorithm scans the data and find the patterns that may represent similar activities and their variations. The algorithm then reports the best patterns that were found and the sensor event data can be compressed using the best pattern. The process is repeated several times until no new patterns can be found that compress the data. The search is carried out using Minimum Description Length principle \cite{Rissanen:1989:SCS:534258} and final set of discovered patterns are clustered using quality threshold clustering method \cite{heyer1999exploring} in which the final number of clusters are not need to be specified apriori. This procedure can generate  `Other' or unknown activity cluster; however, it may contain some already discovered activities. The AD algorithm is performed on the `Other' cluster and already discovered activities can be separated out from it, thereby reducing the `Other' cluster size and the overall recognition of the system reportedly improves. Luca et al. \cite{luca2014detecting} present a method for detecting rare hypermotor seizures in children by attaching accelerometer on the body. They use all the observed data to learn normal behaviour and use extreme value theory to detect deviations from this model. 

The above survey on abnormal activity detection using wearable and vision sensors that can be adapted for fall detection is presented in tabular form in Table \ref{tab:study3}.

\begin{table}[!ht]
\centering
\begin{tabular}
{|l|l|l|}\hline
\textbf{Type of Sensor} & \textbf{Authors}  & \textbf{Technique Used} \\ \hline
\multirow{4}{*}{Wearable Device}            & Yin et al. \cite{DBLP:journals/tkde/YinYP08}  & OSVM, Kernel Nonlinear Regression  \\ \cline{2-3} 
                                            & Quinn et al. \cite{quinn} & SLDS \\ \cline{2-3}
                                            & Cook et al. \cite{cook2012act} & Clustering \\ \cline{2-3}
                                            & Luca et al. \cite{luca2014detecting} & Extreme Value Theory \\ \hline

\multirow{8}{*}{Video}                      & Xiang and Gong \cite{DBLP:conf/iccv/XiangG05a} & Dynamic Bayesian Network \\ \cline{2-3}
                                            & Duong et al. \cite{DBLP:journals/ai/DuongPBV09} & Switching HMM \\ \cline{2-3}
                                            & Zhang et al. \cite{DBLP:conf/cvpr/ZhangGBM05} & Semi-supervised HMM \\ \cline{2-3}
                                            & Jiang et al. \cite{ieee:Jiang:4379786} & HMM \\ \cline{2-3}
                                            & Pruteanu-Malinici and Carin \cite{DBLP:journals/tip/Pruteanu-MaliniciC08} & Infinite HMM \\ \cline{2-3}
                                            & Zhang et al. \cite{DBLP:conf/pakdd/ZhangLGH09} & HDP-HMM, Ensemble Learning  \\ \cline{2-3}
                                            & Hu et al. \cite{DBLP:conf/ijcai/HuZYZY09} & HDP-HMM, OSVM \\ \cline{2-3}
                                            & Antonakaki et al. \cite{DBLP:journals/sigpro/AntonakakiKP09} & HMM, OSVM \\ \cline{2-3}
                                            & Matilainen et al. \cite{DBLP:conf/acivs/MatilainenBS09} & HMM, GMM \\ \cline{2-3}
                                            & Mahajan et al. \cite{mahajan04} & FSM \\ \cline{2-3}
                                            & Parisi and Wermter \cite{Parisi2013Hierarchical} & SOM \\ \hline
\end{tabular}
   \caption{Summary of Studies on Anomaly Detection not Specifically Focused on Fall Detection}
   \label{tab:study3}
  \end{table}

From Tables \ref{tab:study2} and \ref{tab:study3}, we observe that for detecting a fall as an anomaly or adapting abnormal activity detection methods to identify falls, the most common techniques are based on OSVM, OCNN, HMM, GMM and their variants. Some other techniques are also used such as Extreme Value Theory, Clustering, Dynamic Bayesian Network and SLDS. There is a wide range of one-class classification algorithms \cite{Khan:KER:2014} that could be potentially used for detecting falls as anomaly. 

\section{Future Research Directions and Challenges}
\label{sec:conc}
In this paper, we shift our research focus from traditional approaches of fall detection that assume an overtly optimistic assumption that data for falls is sufficiently available.  We discussed that a fall is an abnormal activity that occurs rarely; hence, its data is scarce. This idea is also supported by several recent studies and statistics available from nursing homes \cite{Stone2015Fall, Debard2012Camera, website:cdc}. In these situations with highly skewed or no data for falls, machine learning approaches other than traditional supervised classification must be investigated. Keeping this view, we presented a taxonomy for the study of fall detection that is based on the availability of data and reviewed literature that follows the taxonomy. We identify that considering a fall as an abnormal activity is an important research area, because this reflects the real situation of the world where falls happen infrequently. In the paper, we pointed to some recent studies where a few real fall data is collected over a period of a year or more. This type of experimental set up requires a long term facility, numerous computing resources, rigorous ethics clearances, extensive time and money investment, data analysis and data labelling effort after the experiment is finished. Under these circumstances, identifying falls as abnormal activities makes more sense; however, the overall concept of normal activity needs to be defined and a lot more data may be needed in comparison to the typical supervised classification. 

It has been shown that personalized fall detection systems work better than a general fall detection system that is trained with data from one (or more) person(s) and be adapted to other persons \cite{medrano2016effect}. The generalization of results is problematic in the supervised classification of falls and further aggravates when falls are detected as anomalies by learning models using only data from the normal activities. It is very hard to develop personalized fall detection solution for each and every faller; therefore, research is needed to improve the generalization capabilities of fall detection systems across different persons.

There are several open research areas for fall detection techniques that may require few or no training data for falls.  Auto-encoders can be trained directly from the sensor or vision data and can be used to identify anomalous activities based on their reconstruction error \cite{japkowicz1995novelty, Marchi2015novel}. With Auto-encoders, a reduced set of features can be learned from the raw sensor data that can be used to train other standard one-class classifiers for detecting unseen falls as abnormal activities. 
Recurrent neural networks (RNN) are well-suited for this problem because the sensor data fror human activities is sequential in nature. RNN have been used for intrusion detection \cite{anyanwu2010scalable} and we expect further research to investigate their suitability for unseen fall detection to be very useful. Extreme Value Theory (EVT) has been successfully used to identify rare health conditions among patients in an unsupervised manner \cite{luca2014detecting, clifton2014extending}. The methods based on EVT are mathematically sound to exclusively model rare events that occur in the tails of a distribution and have the potential to be adapated for the fall detection problem. 
The decision theoretic methods for fall detection in the absence of training data (e.g. Khan and Hoey \cite{khan2016dtfall}) assume that agents make rational choices under uncertainty and risks. When humans are involved in decision making under similar circumstances, those decisions are not rational anymore because humans over-estimate the probability of rare events and under-estimate the probability of more-likely events \cite{kahneman1979prospect}. This is the outcome of Prospect Theory and we believe that research in that direction can help in understanding the value of deployment of fall detection devices and their subsequent evaluation by humans. 

A major unresolved question in fall detection methods is the incorporation of costs of errors; however, such costs are not well understood and are hard to compute. This cost also depends on the optimization of the decision threshold to give an optimal choice between false alarms and missed alarms. Since the data for falls is not available; parameter optimization is very challenging in OCC approaches \cite{liu2016modular}. Some approaches reject noisy data from the target class (or normal activities) and use them as validation set for estimating the parameters of the unseen class; however, they may be sensitive to false alarms \cite{khan2014iwaal}. Research efforts are required in this direction to find good strategies to optimize parameters in the absence of no or few data from abnormal activities (or falls in this context). The traditional way to reduce false alarms in a fall detection application is the use of domain knowledge and heuristic rules that are static and hard to generalize. Various researchers have identified different fall risks factors, such as variability in voluntary movement paths of older adults \cite{kearns2012path} as an independent predictor of fall risk. We believe that combining these fall risk scores with the probabilities (or scores) for the unseen falls will be helpful in reducing false alarms. Using the OCC strategies along with combining activity data from multiple data sources such as sensors, video camera, microphone and RFID can result in robust estimation of normal activities, from which falls may be identified effectively as abnormal events. 

We also noted that there is a lack of standard framework to evaluate different fall detection methods and for data repositories. Researchers collect their own activity data in specific settings that may be hard to replicate in other settings. It has been found that the performance of classifiers decrease significantly when they are tested on a dataset different from the one used for training \cite{igual2015comparison}. As discussed earlier, for learning the concept of normal activities, a lot of data is required that should be collected across different persons, age groups and experimental settings. This type of variation may not be present in individual studies and the dataset created from them. An interesting idea would be to combine these publicly available datasets to understand the overall concept of normal activities and perform further research on identifying unseen falls.  

\section*{Declaration}
  Conflicts of Interest : None
  
  Funding: None
  
  Ethical Approval : Not required

\section*{References}
\bibliography{references}
\end{document}